\documentclass{article}
\newcommand{\ignore}[1]{}

\usepackage{microtype}
\usepackage{graphicx}
\usepackage{subcaption}
\usepackage{booktabs} % for professional tables
\usepackage{amsmath}
\usepackage{amssymb}
\usepackage{notations}
\usepackage{natbib}

\usepackage{xcolor}
\definecolor{darkblue}{rgb}{0,0,0.5}
\definecolor{firebrick}{rgb}{0.75,0.125,0.125}
\definecolor{darkgreen}{rgb}{0,0.5,0}
\definecolor{light-gray}{gray}{0.5}
\usepackage[colorlinks=true,linkcolor=firebrick,citecolor=darkgreen,urlcolor=darkblue]{hyperref}

\usepackage[accepted]{icml2018}

\icmltitlerunning{Spurious samples in deep generative models: bug or feature?}

\begin{document}

\twocolumn[
\icmltitle{Spurious samples in deep generative models: bug or feature?}



\icmlsetsymbol{equal}{*}

\begin{icmlauthorlist}
\icmlauthor{Bal\'azs K\'egl}{cnrs}
\icmlauthor{Mehdi Cherti}{mines}
\icmlauthor{Ak{\i}n Kazak\c{c}{\i}}{mines}
\end{icmlauthorlist}

\icmlaffiliation{cnrs}{CNRS/Universit\'{e} Paris-Saclay}
\icmlaffiliation{mines}{PSL Research University, CGS-I3 UMR 9217}
%\icmlcorrespondingauthor{}{balazs.kegl@gmail.com}
%\icmlcorrespondingauthor{Eee Pppp}{ep@eden.co.uk}

\icmlkeywords{Machine Learning, ICML}

\vskip 0.3in
]



\printAffiliationsAndNotice{\icmlEqualContribution} % otherwise use the standard text.

\begin{abstract}
Traditional wisdom in generative modeling literature is that spurious samples that a model can generate are errors and they should be avoided. Recent research, however, has shown interest in studying or even exploiting such samples instead of eliminating them. In this paper, we ask the question whether such samples can be eliminated all together without sacrificing coverage of the generating distribution. For the class of models we consider, we experimentally demonstrate that this is not possible without losing the ability to model some of the test samples. While our results need to be confirmed on a broader set of model families, these initial findings provide partial evidence that spurious samples share structural properties with the learned dataset, which, in turn, suggests they are not simply errors but a feature of deep generative nets.
\end{abstract}

\section{Introduction}

The goal of unsupervised modelling is to learn a characterization of the data generating distribution from a set of training instances. Generative modelling also aims at a constructive procedure to generate samples from the learned distribution. Evaluating the quality of these models is not trivial:~\citep{theis2015note} shows that most commonly used criteria such as average log-likelihood, Parzen window estimates, and visual
quality of samples are largely independent of each other when the data is
high-dimensional. More importantly, they conclude that extrapolation from one criterion to another is not warranted and generative models need to be evaluated directly with respect to the application(s) they were intended for~~\cite{theis2015note}. 

\begin{figure}[!ht]
\centering
\begin{subfigure}[b]{\columnwidth}
\includegraphics[width=\columnwidth]{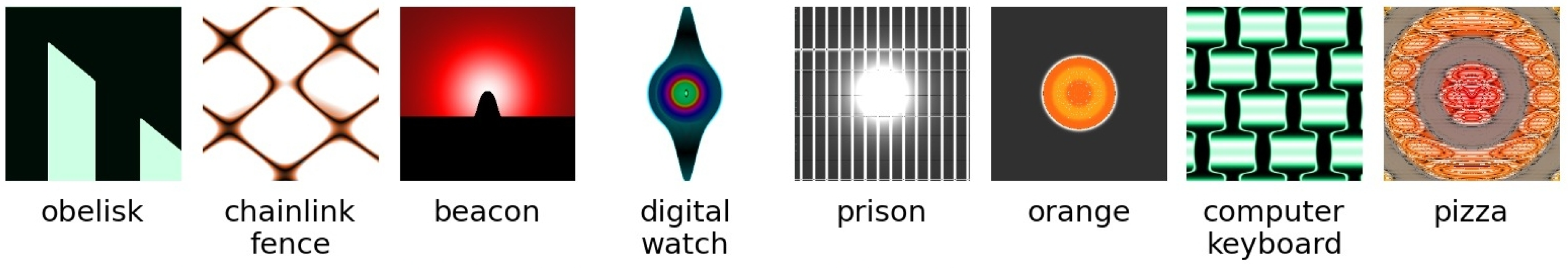}
\caption{``Synthetic'' objects from imagenet categories from Figure~7 of \cite{nguyen2015innovation}\label{figInnovationEngine}}
\end{subfigure}

\begin{subfigure}[b]{\columnwidth}
\includegraphics[width=0.24\columnwidth]{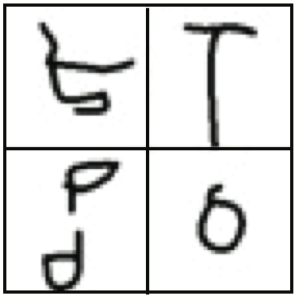}
\includegraphics[width=0.24\columnwidth]{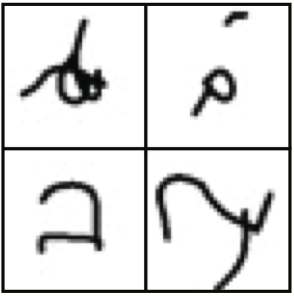}
\includegraphics[width=0.24\columnwidth]{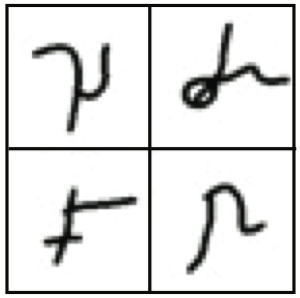}
\includegraphics[width=0.24\columnwidth]{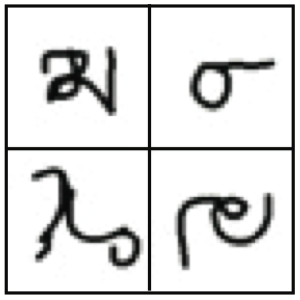}
\caption{``Unconstrained'' symbols from Figure~7 of \cite{lake2015human}\label{figOneShot}}
\end{subfigure}

\begin{subfigure}[b]{\columnwidth}
\includegraphics[width=\columnwidth]{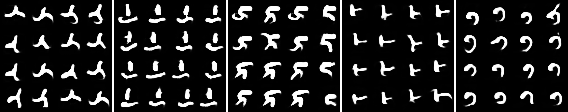}
\caption{New types of symbols from Figure~6 of \cite{kazakcci2016digits}\label{figNewTypes}}
\end{subfigure}

\begin{subfigure}[b]{0.5\columnwidth}
\includegraphics[width=\columnwidth]{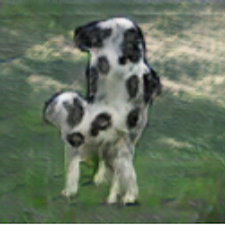}
\caption{
Non-recognizable animal generated by~\cite{goodfellow2016nips}.
}
\end{subfigure}

\caption{Examples of spurious objects.}
\label{figOutOfDistributionExamples}
\end{figure}

The difficulties in the evaluation of generative models become exacerbated when we consider the notion of \emph{spurious} samples~\cite{bengio2013generalized}. These kinds of examples (Figure~\ref{figOutOfDistributionExamples}) are ubiquitous in the mainstream generative model  literature. Traditionally, researchers strive to eliminate these samples~\cite{bengio2013generalized,goodfellow2016nips,salimans2016improved} since they are considered failures. There is also a chance that in published work such samples are underreported. By contrast, recent work \cite{nguyen2015innovation, lake2015human, kazakcci2016digits, cherti2017out} has shown  growing interest in exploiting or studying these kind of samples. \citet{kazakcci2016digits} 
found that it was quite easy to generate examples that had zero likelihood under any possible notion of likelihood; more precisely, they generated symbols by models trained on digits which were not digits under any notion of what a digit is (Figure~\ref{figNewTypes}). \citet{lake2015human} called these examples ``unconstrained''. Such studies highlight that spurious samples should not be discarded (e.g., as ``noise''): they share deep structural properties with examples of the training set, yet they are obviously not coming from the distribution that generated these training sets.

\begin{figure}[!ht]
\centering
\begin{subfigure}{0.24\columnwidth}
\includegraphics[width=\columnwidth]{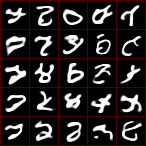}
\caption{}\label{mnist_spurious}
\end{subfigure}
\begin{subfigure}{0.24\columnwidth}
\includegraphics[width=\columnwidth]{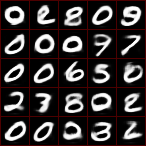}
\caption{}\label{mnist_missing_modes}
\end{subfigure}
\begin{subfigure}{0.24\columnwidth}
\includegraphics[width=\columnwidth]{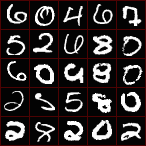}
\caption{}\label{mnist_tail}
\end{subfigure}
\caption{Generated hand-written symbols illustrating (a) spurious modes and (b) missing modes. Detecting missing modes visually is hard; even in case of the simple domain of MNIST we need a reference sample (c) from the tail of the distribution to notice that the model that generated (b) is probably missing some of the modes.}
\label{figGeneratedSymbols}
\end{figure}

Given these shared structural properties between relevant and spurious samples, one trivial and fundamental question needs to be answered: is it possible to get rid of all spurious samples without sacrificing the coverage of a model? This question translates into understanding the relationship between the following kinds of failure modes:
\begin{enumerate}
  \item \emph{Spurious modes}: whether the model generates objects that clearly do not belong to the domain (Figure~\ref{mnist_spurious}).
  \item \emph{Missing modes} or the lack of \emph{coverage}: whether the model can generate \emph{all} objects (e.g., all bedrooms or handwritten digits) of the domain (Figure~\ref{mnist_missing_modes}).
\end{enumerate}

In other words, in this paper we ask the question whether it is possible at all to learn a training set (e.g., MNIST digits) and \emph{only} the training set, or when we learn a representation of all the training set, we will inescapably learn a larger set of structurally similar objects, not present in the training set. This would mean that for achieving full coverage, we need to live with spurious objects. On the specific set of models we studied, the answer seems affirmative: we either discard a subset of digits, learning only the bulk of the distribution, losing the tail, or if we pick up the full set of digits, the models also naturally represent a much larger set of structurally similar objects.

Contrary to latest research on generative modeling, we chose MNIST as the main data set of study, complemented by the HWRT dataset \cite{thoma2015line} of handwritten mathematical symbols for out-of-class objects, mainly since the experimental setup required to train a large number of generative models. Unlike positive results of learnability (a new algorithm which \emph{can} learn MNIST), our argument is not harmed by this limitation. Essentially, we provide evidence that \emph{even a simple distribution or training set} is difficult to learn properly. This is a negative result; eliminating spurious modes while covering the full distribution should be \emph{more difficult} on larger, more heterogeneous and higher dimensional data sets.

The paper has the following contributions:
\begin{itemize}
  \item We provide an experimental framework to study and to quantify both spurious and missing modes in a specific type of generative models.
  \item We define a new metric that can be used to tune the spurious/missing mode trade-off and for selecting models that achieve the best compromise.
   \item We show that, for the type of models we studied, it is impossible to, at the same time, eliminate spurious modes while learning all genuine modes.
%   \item We study both the set of spurious modes we detect and the missing modes and show that they are not random ...
\end{itemize}

\section{Spurious samples and the evaluation of generative models }

\citet{theis2015note} argues against using simple Parzen-based~\citep{breuleux2009unlearning}) likelihood estimators, even going as far as saying that if the goal is visual appeal, then likelihood itself is not necessarily a good measure. Detecting spurious modes (Figure~\ref{mnist_spurious}; Figure~\ref{figOutOfDistributionExamples}) visually is relatively easy. So, when the goal is not to generate these spurious examples \cite{bengio2013generalized} or failures \cite{salimans2016improved}, non-likelihood-based metrics concentrate on eliminating them. Inception (or, more generally, \emph{objectness}) score \cite{salimans2016improved} and Frechet (inception) distance \cite{heusel2017gans} require a sub-class predictor to label sub-class modes (e.g., imagenet or MNIST classes), somewhat defeating the very goal of unsupervised learning. Moreover, by design, they are susceptible to missing modes that were not labeled, within the classes used in the prediction task. For example, Figure~\ref{mnist_missing_modes} displays a generated sample which is missing some labels (sub-classes, from a point of view of modeling digits) but it also shows less variety within the classes. Missing classes are detected by objectness but the lack of within-class variety is not. One may argue that objectness even \emph{penalizes} tail examples since the predictor is better at classifying typical examples than tail examples. Also note that detecting missing modes visually is hard, even in case of the simple domain of MNIST we need a reference sample (Figure~\ref{mnist_tail}) from the tail of the test sample to notice that the model that generated the sample in Figure~\ref{mnist_missing_modes} is probably missing some of the modes.

None of these metrics seem suitable to analyze the relationship between spurious and missing modes. In the next section, we propose a new metric $\Delta$ that we derive from the in-class and out-of-class reconstruction rates. As we will see, this metric has the advantage that it does not require a sub-class classifier (like the inception imagenet classifier). On the other hand, it requires a control sample on which we can measure out-of-class reconstruction rates. In practical situations, say, in a data challenge, the control set can be kept hidden from the modelers, making it less likely that they overfit $\Delta$. It is even possible to use several proxy control sets and to combine the resulting $\Delta$ scores using various statistics (e.g., mean or min).

Besides the control set, another requirement for applying l is that each trained model $M$ have to be able to answer to the binary question whether a given object $x$ can or cannot be reconstructed by $M$. Autoencoders~\citep{vincent2010stacked, bengio2013generalized} and autoregressive models \citep{oord2016pixel,van2016conditional} have this property but, for example, GANs\cite{goodfellow2014generative} do not.

\section{The formal setup}

\begin{figure}[!ht]
\centering
\includegraphics[width=1.0\columnwidth]{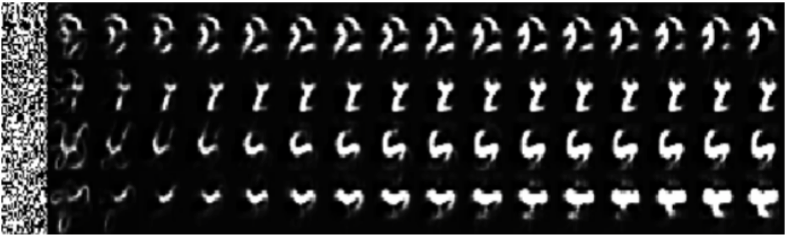}
\caption{Iterative generation with autoencoders. We start from random images obtained by randomly picking pixel intensities uniformly from 0 to 1, and we repeatedly apply the autoencoder on the images for several iterations. Each row corresponds to an independent sample, while columns correspond to iterations. The images of the last iteration correspond to approximate fixed points of the autoencoder. In other words, those points $x$ can be reconstructed very well by the autoencoder, that is $M(x) \approx x$.}
\label{figIterativeGeneration}
\end{figure}

\begin{figure}[!ht]
\centering
\begin{subfigure}[b]{0.9\columnwidth}
\includegraphics[width=0.45\columnwidth]{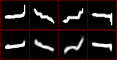}
\includegraphics[width=0.45\columnwidth]{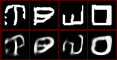}
\caption{$\theta=30$}
\end{subfigure}

\begin{subfigure}[b]{0.9\columnwidth}
\includegraphics[width=0.45\columnwidth]{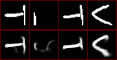}
\includegraphics[width=0.45\columnwidth]{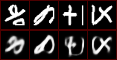}
\caption{$\theta=50$}
\end{subfigure}

\begin{subfigure}[b]{0.9\columnwidth}
\includegraphics[width=0.45\columnwidth]{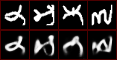}
\includegraphics[width=0.45\columnwidth]{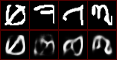}
\caption{$\theta=80$}
\end{subfigure}

\caption{The effect of the threshold $\theta$ on images that are considered as``recognized'' and the ones that are not. For each value of $\theta$, the left panels correspond to images considered as recognized by the model, while the right panels correspond to images considered that are not. Within each panel, the top row contains the original images and the bottom row contains their corresponding reconstructions.
The reconstructions are obtained from a convolutional autoencoder trained on MNIST training set and the reconstructions are obtained from the test set of handwritten mathematical symbols.
}
\label{figTheta}
\end{figure}

Let $\cX$ be the set of all images of dimension $28 \times 28$ with gray-scale pixel values in $[0, 1]$. Each model $M$ is an autoencoder that represents a manifold $\cM_M = \{x \in \cX: M(x) \approx x\} = \{x \in \cX: \|M(x) - x\|_1 < \theta$\}. We shall also say that $M$ can \emph{reconstruct} elements of $\cM_M$ or $\cM_M$ itself. The threshold $\theta$ was set to $50$ experimentally in order to maximize the dynamic range of our scores (smaller or larger thresholds would have resulted in models that reconstructed few digits and symbols or most of them, respectively, see Figure~\ref{figTheta}). The manifold $\cM_M$ is loosely related to the distribution of the images \cite{alain2014regularized}, more precisely, it is the approximate \emph{support} of the distribution when the autoencoder is used in an iterative generative mode targeting its fixed points (Figure~\ref{figIterativeGeneration}). In our setup we are not interested in the actual likelihood assigned to the fixed point (related to the measure of the set of random seeds that generate the fixed point), rather to a yes/no answer to the question whether the model $M$ can represent/generate a given image $x$. Also note that the actual setup of turning the generative model into an oracle that can answer to the question ``can you reconstruct $x$?'' may vary, depending on the model. The particular setup is somewhat independent of the metrics we propose.

The models are all trained on the 60000 training images of MNIST. We use the test set $\cD$ of MNIST for evaluation. For detecting spurious modes, we also use a control set of handwritten mathematical symbols from~\cite{thoma2015line}. This dataset is originally vectorized and consists in a sequence of coordinates $(x, y)$ for each example. We rasterized it by joining the coordinates by segments with a thickness similar to MNIST. We also padded the images with zeros as it was done in MNIST. The full dataset has 369 classes of mathematical symbols. We remove all the classes with less than 100 examples, obtaining a total of 343 classes and 151853 examples. We randomly split the full set into a training set of 60000 examples and a test set $\cS$ of 91853 examples. The training sets were used to train a digit vs. symbol classifier which was used in the analysis in Section~\ref{secTradeOff}. We denote the posterior probabilities output by this classifier by $p(D | x)$ and $p(S | x)$.

%to find the set of digits $\cD_\cS$ that look like (cannot be differentiated from) symbols and the set of symbols $\cS_\cD$ that look like digits. .

Let $\cD_M = \cM_M \cap \cD$ and $\cS_M = \cM_M \cap \cS$ be the set of digits and symbols, respectively, which a trained model $M$ can reconstruct. The \emph{in-class reconstruction rate} (IRR) and \emph{out-of-class reconstruction rate} (ORR) of a model $M$ are defined as
\begin{equation}\label{eqnIRR}
  \text{IRR}(M) = \frac{|\cM_M \cap \cD|}{|\cD|} = \frac{|\cD_M|}{|\cD|}
\end{equation}
and
\begin{equation}\label{eqnORR}
  \text{ORR}(M) = \frac{|\cM_M \cap \cS|}{|\cS|} = \frac{|\cS_M|}{|\cS|}.
\end{equation}
For each model $M$, we will use $1 - IRR(M)$ and $ORR(M)$ to quantify the ``measure'' of the missing modes and spurious modes, respectively. Assuming that the $\cD$ and $\cS$ are generated i.i.d, $1 - IRR(M)$ and $ORR(M)$ are unbiased estimates of probabilities that a digit $x \not\in \cM_M$ and a symbol $x \in \cM_M$, respectively. Thus, $1 - IRR(M)$ is indeed a measure of the missing modes under the sampling distribution of $\cD$, but $ORR(M)$ is only a proxy of the measure of the spurious modes since it only covers those modes that are sampled by the symbol set $\cS$. In the experiments we will use the difference
\begin{equation}\label{eqnDelta}
 \Delta(M) = \text{IRR}(M) - \text{ORR}(M)
\end{equation}
as a proxy metrics for identifying ``good'' models, that is, good compromises of low rates of both spurious and missing modes.

%  $\cD_\cD = \cD \backslash \cD_\cS$, $\cS_\cS = \cS \backslash \cS_\cD$.
% \begin{equation}
%   \text{ORR}^\prime(M) = \frac{|\cM_M \cap \cS_\cD|}{|\cS|}
% \end{equation}

\section{Experiments}

In the experiments, we use a family of convolutional autoencoders. All the models consists in a set of $L$ convolutional layers on the encoder, followed by a set of $L$ convolutional layers with padding~\citep{dumoulin2016guide} (to increase the size of the feature maps) on the decoder, thus a total of $2L$ layers. Each convolutional layer has a filter of size $k$ and use the \emph{ReLU} activation function. We apply an activation function $\text{act}(\cdot)$ in the bottleneck ("code").
The output layer used a sigmoid activation function.

To explore the space of the architectures, we vary several hyperparameters, while we fix others. We vary the number of layers $L$ from 1 to 6.
We use 128 feature maps in all the layers, except in the bottleneck layer where the number of feature maps varies and can take values from $\{2, 4, 8, 16, 32, 64, 128\}$. We use a filter size of $k=5$ in all the layers and a stride of $1$. For the activation function of the bottleneck $\text{act}(\cdot)$, we use the spatial Winner-Take-All (WTA) activation used in ~\cite{makhzani2015winner}. 
In each feature map, $\text{spatialwta}$ zeroes out all the activations except the activation with the maximum value, thus backpropagating only through through the activation with the maximum value in each feature map. After applying $\text{spatialwta}$, we apply an additional sparsity activation, which we call $\text{channelwta}$. $\text{channelwta}$ is parametrized by a sparsity rate $\rho$, it keeps only the feature maps with the highest activation\footnote{Only a single activation is greater than zero per feature map after having applied $\text{spatialwta}$}, zeroing out $\rho\%$ of them to achieve a channel-wise sparsity rate of $\rho$. We use the values $\rho \in \{0, 0.1, 0.2, 0.4, 0.5, 0.7, 0.9\}$, where $\rho = 0$ means we do not use $\text{channelwta}$. We also use denoising~\citep{vincent2010stacked} with 
 \emph{salt and pepper} noise by varying probabilities of corruption $p_{\text{corruption}}$ where $p_{\text{corruption}} \in  \{0, 0.1, 0.2, 0.3, 0.4, 0.5\}$.
All the models are trained on the MNIST training dataset with the reconstruction error objective, using mean squared error (MSE). A total of 187 models have been trained for the current experiments.

\subsection{The spurious/missing mode trade-off}
\label{secTradeOff}

\begin{figure*}[!ht]
\centering
\includegraphics[width=0.8\textwidth]{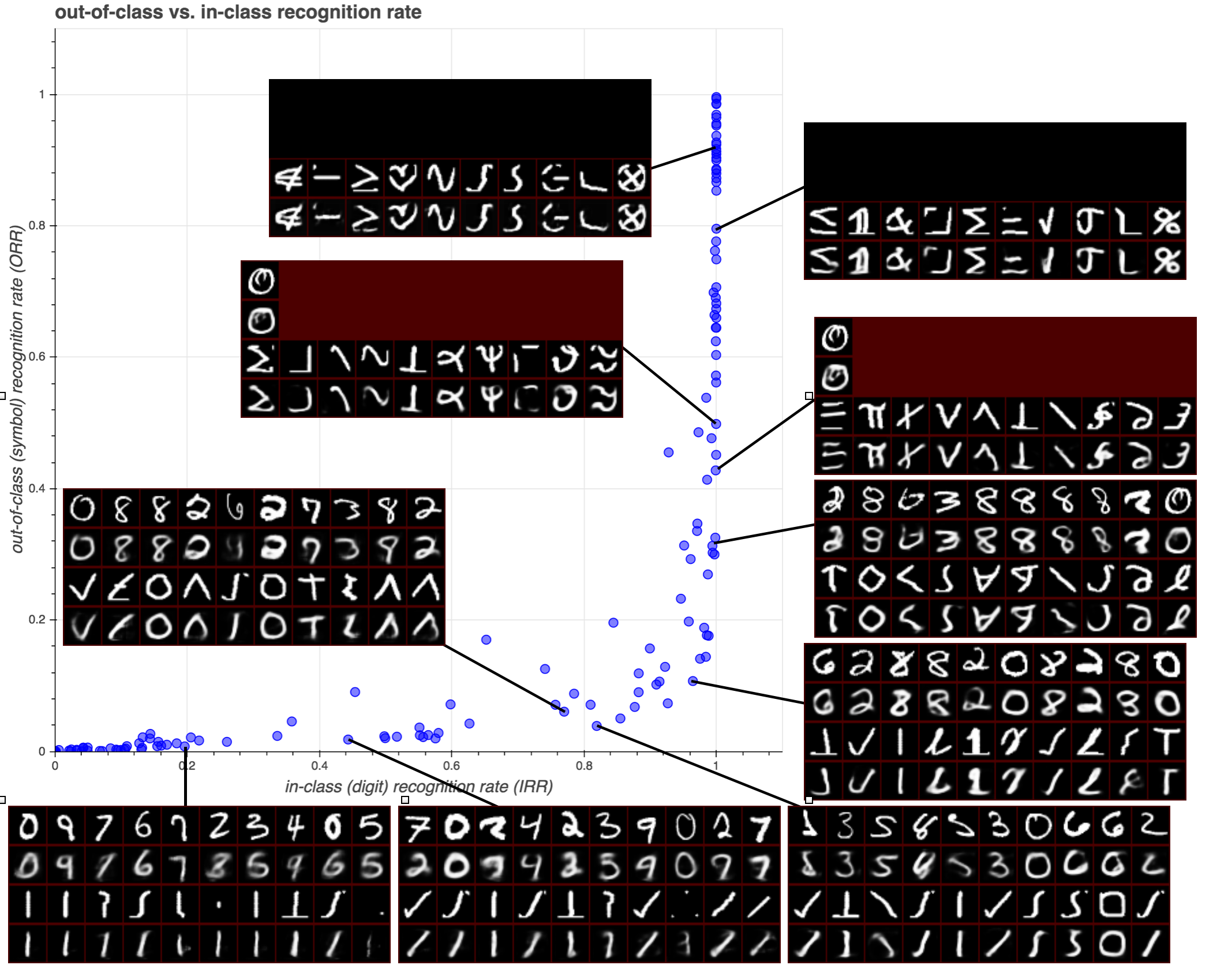}
\caption{The out-of-class recognition rate ORR (\ref{eqnORR}) versus the in-class recognition rate IRR (\ref{eqnIRR}). The panels show a random set of digits that cannot be reconstructed by the selected model in the first row, and their (attempted) reconstructions in the second row. These images represent missing modes. The third row of each panel is the symbols that the selected model can reconstruct, with the reconstruction in the fourth row. These images represent spurious modes. 
An interactive version of the plot, where the reader can click on any dot to see the corresponding panel is available at \url{https://goo.gl/ehbrb3}.
}
\label{figORRVsIRRModes}
\end{figure*}

Figure~\ref{figORRVsIRRModes} shows the out-of-class recognition rate ORR (\ref{eqnORR}) versus the in-class recognition rate IRR (\ref{eqnIRR}) of each trained model $M$. For selected models, we show some digits $x \in \cD \backslash \cD_M$ that they cannot reconstruct, illustrating the missing modes, and some symbols $x \in \cS_M$ that they can, illustrating the spurious modes. The most important observation here is that none of the models are perfect: they either reconstruct all digits but also a large portion of the symbols, or they have a low rate of spurious modes but missing also a large portion of the digits. While the actual numbers are somewhat dependent on the reconstruction threshold $\theta$, with the threshold we selected $\theta = 50$, the first model that can reconstruct $99.5\%$ of the digits can also reconstruct about $30\%$ of the symbols, and the first model that discards $95\%$ of the symbols can only reconstruct about $60\%$ of the digits. The model with the best $\Delta = 0.85$ (\ref{eqnDelta}) makes a compromise of reconstructing $95\%$ of the digits and about $10\%$ of the symbols.

The panels attached to selected models $M$ show both the original digits and symbols (first and third rows) $x$ and the reconstructed digits and symbols (second and fourth rows). As we move from $(0,0)$ towards $(1, 1)$, missing modes become more ant more ``esoteric'' until they disappear completely, while spurious modes become richer and richer as models pick up more and more symbols. 

One criticism of the methodology could be that the set of symbols and digits overlap, and the reconstructed symbols $\cS_M$ all look like digits (coming from $\cD \cap \cS$). It is clear from the examples that this is not the case: most reconstructed symbols do not look like digits to a human evaluator. To make this counterargument more formal, we trained a digit vs. symbol classifier $p(S|x)$. The low test error of $0.2\%$ showed that indeed most symbols can be recognized as symbols by an ``objective'' classifier. There still remained a doubt on whether the models ``pulled'' all the reconstructed symbols $M(x)$ into the digit set, so we also looked at the symbol classification rate in $\{M(x): x \in \cS_M\}$, that is, the rate of reconstructed symbols that looked like digits to the discriminator. While the rates were higher than $0.2\%$, they remained in the low 10s, confirming that indeed, most reconstructed symbols are spurious, even under this more stringent criterion.

\begin{figure*}[!ht]
\centering
\includegraphics[width=0.8\textwidth]{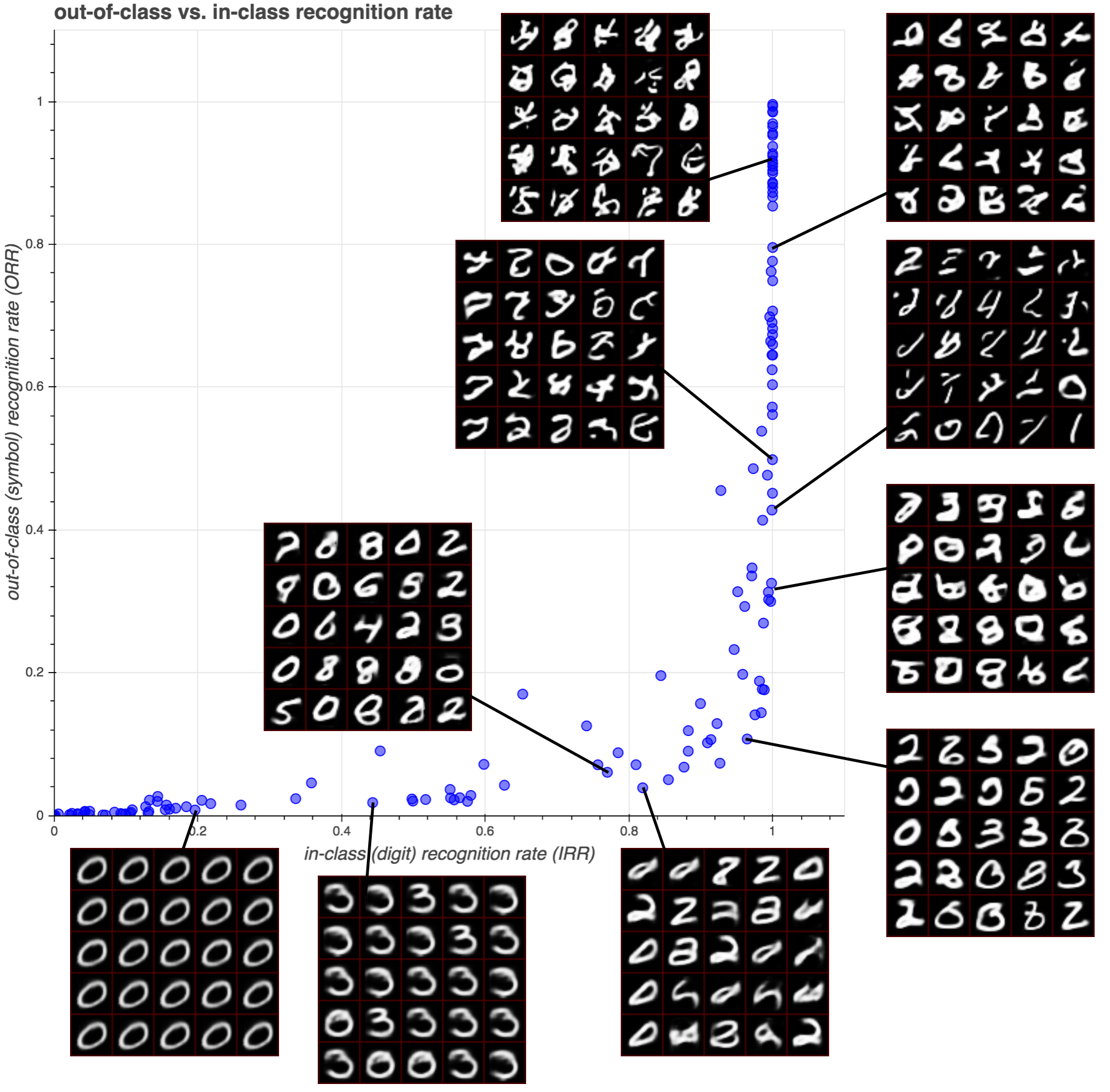}
\caption{The out-of-class recognition rate ORR (\ref{eqnORR}) versus the in-class recognition rate IRR (\ref{eqnIRR}). The panels show a random set of images generated from random seeds by the procedure described in Figure~\ref{figIterativeGeneration} for selected models.
An interactive version of the plot, where the reader can click on any dot to see the corresponding panel is available at \url{https://goo.gl/Ltzp3V}.
}
\label{figORRVsIRRSamples}
\end{figure*}

Figure~\ref{figORRVsIRRSamples} shows the same ORR vs. IRR plot, but panels of selected models show images generated from random seeds using the procedure in Figure~\ref{figIterativeGeneration}. Models towards $(0,0)$ generate overwhelmingly digits, but the variability of these digits is visibly lower than in MNIST. Models towards $(1,1)$ generate overwhelmingly spurious symbols. These models are typical candidates for research in novelty generation \cite{nguyen2015innovation,lake2015human,kazakcci2016digits,cherti2017out}. Finally, models towards $(0,1)$ are those that could be considered a good compromise, generating mostly digit-looking symbols with a high variability.

Using $\Delta(M)$ can be considered as a metric for selecting these models, and the full IRR-ORR plane can be used to tune the trade-off between accepting either spurious or missing modes. Note also that in practical situations, say, in a data challenge, the control set $\cS$ can be kept hidden from the modelers, making it less likely that they overfit the particular ORR metrics and thus $\Delta$. It is even possible to use several proxy control sets and to combine the resulting $\Delta$ scores using various statistics (e.g., mean or min). 

\subsection{Comparing $\Delta$ to objectness}

\begin{figure*}[!ht]
\centering
\includegraphics[width=0.8\textwidth]{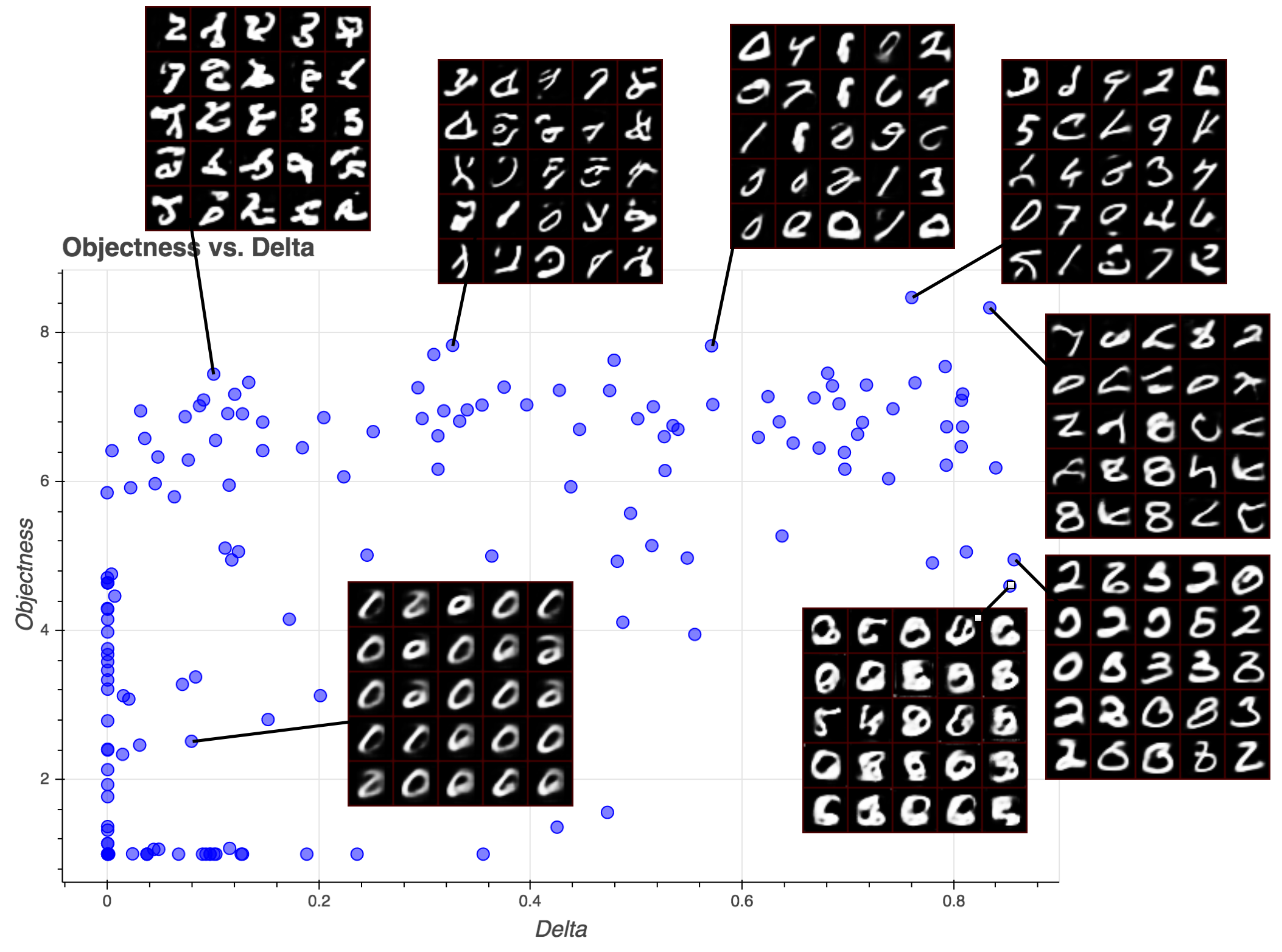}
\caption{Objectness \cite{salimans2016improved} vs. $\Delta$, visualizing generated images of selected models.
An interactive version of the plot, where the reader can click on any dot to see the corresponding panel is available at \url{https://goo.gl/KWuTvk}.
}
\label{figObjectness}
\end{figure*}

Objectness (or inception score) \cite{salimans2016improved} is one of the popular non-likelihood-based quality metrics. It requires a sub-class classifier so we trained for it a standard convnet for classifying MNIST digits. Figures~\ref{figObjectness} shows the the scatterplot of objectness vs. $\Delta$. The two metrics agree on what bad models are, but not on what good models are. Furthermore, it is hard to say if there is any correlation between these measures and human judgement. Objectness tends to be insensitive to spurious modes, possibly because of the ``blind spots'' of the classifier (it confidently classifies spurious symbols into one of the digit classes). 

\section{Discussion}
The paper presents an investigation of the spurious samples in deep generative models and their relationship with a model's ability to effectively learn the domain being modelled. Through a set of experiments and for a specific model family, we have shown that there is a trade-off between a model's potential to generate spurious samples and its effectiveness for covering all the available training instances. 
This implies that, at least for the models we considered, one cannot eliminate spurious samples without sacrificing the model's ability to generate some data we actually want to model. The metrics we used in this study, in-class and out-of-class reconstruction rates and their difference, can be used as an alternative non-likelihood-based metrics to tune the spurious/missing mode trade-off and for selecting models that achieve the best compromise.

\bibliographystyle{icml2018}
\bibliography{main}

\begin{thebibliography}{18}
\providecommand{\natexlab}[1]{#1}
\providecommand{\url}[1]{\texttt{#1}}
\expandafter\ifx\csname urlstyle\endcsname\relax
  \providecommand{\doi}[1]{doi: #1}\else
  \providecommand{\doi}{doi: \begingroup \urlstyle{rm}\Url}\fi

\bibitem[Alain \& Bengio(2014)Alain and Bengio]{alain2014regularized}
Alain, Guillaume and Bengio, Yoshua.
\newblock What regularized auto-encoders learn from the data-generating
  distribution.
\newblock \emph{The Journal of Machine Learning Research}, 15\penalty0
  (1):\penalty0 3563--3593, 2014.

\bibitem[Bengio et~al.(2013)Bengio, Yao, Alain, and
  Vincent]{bengio2013generalized}
Bengio, Yoshua, Yao, Li, Alain, Guillaume, and Vincent, Pascal.
\newblock Generalized denoising auto-encoders as generative models.
\newblock In \emph{Advances in Neural Information Processing Systems}, pp.\
  899--907, 2013.

\bibitem[Breuleux et~al.(2009)Breuleux, Bengio, and
  Vincent]{breuleux2009unlearning}
Breuleux, Olivier, Bengio, Yoshua, and Vincent, Pascal.
\newblock Unlearning for better mixing.
\newblock \emph{Universite de Montreal/DIRO}, 2009.

\bibitem[Cherti et~al.(2017)Cherti, K{\'e}gl, and
  Kazak{\c{c}}{\i}]{cherti2017out}
Cherti, Mehdi, K{\'e}gl, Bal{\'a}zs, and Kazak{\c{c}}{\i}, Ak{\i}n.
\newblock Out-of-class novelty generation: an experimental foundation.
\newblock In \emph{Tools with Artificial Intelligence (ICTAI), 2017 IEEE 29th
  International Conference on Tools with Artificial Intelligence}. IEEE, 2017.

\bibitem[Dumoulin \& Visin(2016)Dumoulin and Visin]{dumoulin2016guide}
Dumoulin, Vincent and Visin, Francesco.
\newblock A guide to convolution arithmetic for deep learning.
\newblock \emph{arXiv preprint arXiv:1603.07285}, 2016.

\bibitem[Goodfellow(2016)]{goodfellow2016nips}
Goodfellow, Ian.
\newblock Nips 2016 tutorial: Generative adversarial networks.
\newblock \emph{arXiv preprint arXiv:1701.00160}, 2016.

\bibitem[Goodfellow et~al.(2014)Goodfellow, Pouget-Abadie, Mirza, Xu,
  Warde-Farley, Ozair, Courville, and Bengio]{goodfellow2014generative}
Goodfellow, Ian, Pouget-Abadie, Jean, Mirza, Mehdi, Xu, Bing, Warde-Farley,
  David, Ozair, Sherjil, Courville, Aaron, and Bengio, Yoshua.
\newblock Generative adversarial nets.
\newblock In \emph{Advances in Neural Information Processing Systems}, pp.\
  2672--2680, 2014.

\bibitem[Heusel et~al.(2017)Heusel, Ramsauer, Unterthiner, Nessler, and
  Hochreiter]{heusel2017gans}
Heusel, Martin, Ramsauer, Hubert, Unterthiner, Thomas, Nessler, Bernhard, and
  Hochreiter, Sepp.
\newblock Gans trained by a two time-scale update rule converge to a local nash
  equilibrium.
\newblock In \emph{Advances in Neural Information Processing Systems}, pp.\
  6629--6640, 2017.

\bibitem[Kazak{\c{c}}{\i} et~al.(2016)Kazak{\c{c}}{\i}, Cherti, and
  K{\'e}gl]{kazakcci2016digits}
Kazak{\c{c}}{\i}, Ak{\i}n, Cherti, Mehdi, and K{\'e}gl, Bal{\'a}zs.
\newblock Digits that are not: Generating new types through deep neural nets.
\newblock In \emph{Proceedings of the Seventh International Conference on
  Computational Creativity}, 2016.

\bibitem[Lake et~al.(2015)Lake, Salakhutdinov, and Tenenbaum]{lake2015human}
Lake, Brenden~M, Salakhutdinov, Ruslan, and Tenenbaum, Joshua~B.
\newblock Human-level concept learning through probabilistic program induction.
\newblock \emph{Science}, 350\penalty0 (6266):\penalty0 1332--1338, 2015.

\bibitem[Makhzani \& Frey(2015)Makhzani and Frey]{makhzani2015winner}
Makhzani, Alireza and Frey, Brendan~J.
\newblock Winner-take-all autoencoders.
\newblock In \emph{Advances in Neural Information Processing Systems}, pp.\
  2791--2799, 2015.

\bibitem[Nguyen et~al.(2015)Nguyen, Yosinski, and Clune]{nguyen2015innovation}
Nguyen, Anh~Mai, Yosinski, Jason, and Clune, Jeff.
\newblock Innovation engines: Automated creativity and improved stochastic
  optimization via deep learning.
\newblock In \emph{Proceedings of the 2015 on Genetic and Evolutionary
  Computation Conference}, pp.\  959--966. ACM, 2015.

\bibitem[Oord et~al.(2016)Oord, Kalchbrenner, and Kavukcuoglu]{oord2016pixel}
Oord, Aaron van~den, Kalchbrenner, Nal, and Kavukcuoglu, Koray.
\newblock Pixel recurrent neural networks.
\newblock \emph{arXiv preprint arXiv:1601.06759}, 2016.

\bibitem[Salimans et~al.(2016)Salimans, Goodfellow, Zaremba, Cheung, Radford,
  and Chen]{salimans2016improved}
Salimans, Tim, Goodfellow, Ian, Zaremba, Wojciech, Cheung, Vicki, Radford,
  Alec, and Chen, Xi.
\newblock Improved techniques for training {GAN}s.
\newblock \emph{arXiv preprint arXiv:1606.03498}, 2016.

\bibitem[Theis et~al.(2015)Theis, Oord, and Bethge]{theis2015note}
Theis, Lucas, Oord, A{\"a}ron van~den, and Bethge, Matthias.
\newblock A note on the evaluation of generative models.
\newblock \emph{arXiv preprint arXiv:1511.01844}, 2015.

\bibitem[Thoma(2015)]{thoma2015line}
Thoma, Martin.
\newblock On-line recognition of handwritten mathematical symbols.
\newblock \emph{arXiv preprint arXiv:1511.09030}, 2015.

\bibitem[van~den Oord et~al.(2016)van~den Oord, Kalchbrenner, Espeholt,
  Vinyals, Graves, et~al.]{van2016conditional}
van~den Oord, Aaron, Kalchbrenner, Nal, Espeholt, Lasse, Vinyals, Oriol,
  Graves, Alex, et~al.
\newblock Conditional image generation with pixelcnn decoders.
\newblock In \emph{Advances in Neural Information Processing Systems}, pp.\
  4790--4798, 2016.

\bibitem[Vincent et~al.(2010)Vincent, Larochelle, Lajoie, Bengio, and
  Manzagol]{vincent2010stacked}
Vincent, Pascal, Larochelle, Hugo, Lajoie, Isabelle, Bengio, Yoshua, and
  Manzagol, Pierre-Antoine.
\newblock Stacked denoising autoencoders: Learning useful representations in a
  deep network with a local denoising criterion.
\newblock \emph{The Journal of Machine Learning Research}, 11:\penalty0
  3371--3408, 2010.

\end{thebibliography}

\end{document}